\documentclass[12pt]{article}
\usepackage[utf8]{inputenc}

\usepackage{amsfonts}
\usepackage[ruled,vlined]{algorithm2e}

\usepackage{fancyhdr}
\usepackage{comment}
\usepackage[a4paper, top=2.4cm, bottom=2.4cm, left=2.2cm, right=2.2cm]%
{geometry}
\usepackage{times}
\usepackage{amsmath}
\usepackage{changepage}
\usepackage{amssymb}
\usepackage{graphicx}%
\usepackage{multirow}
\usepackage{float}

\usepackage{color}
\usepackage{xcolor,colortbl}
\usepackage{hyperref}
\usepackage{setspace} 
\usepackage{listings}

\usepackage{float}
\usepackage{amsmath}
\usepackage[
    backend=biber,
    style=ieee,
  ]{biblatex}
\usepackage{authblk}
\usepackage{setspace}

\DeclareUnicodeCharacter{0301}{\'{e}}

\begin{document}
\doublespacing

\title{That looks interesting! Personalizing Communication and Segmentation with Random Forest Node Embeddings}
\author[1]{Weiwei Wang}
\author[2]{Wiebke Eberhardt}
\author[2,3]{Stefano Bromuri}

\affil[1]{\small Brightlands Institute for Smart Society (BISS) - Maastricht University, Heerlen, The Netherlands}
\affil[2]{\small Center for Actionable Research of the Open University (CAROU), Heerlen, The Netherlands}

\affil[3]{\small Computer Science Department, Open University of the Netherlands, Heerlen, The Netherlands }

\maketitle

\begin{abstract}

Communicating effectively with customers is a challenge for many marketers, but especially in a context that is both pivotal to individual long-term financial well-being and difficult to understand: pensions. Around the world, participants are reluctant to consider their pension in advance, it leads to a lack of preparation of their pension retirement\cite{binswanger2012real,wiener2008framework}. In order to engage participants to obtain information on their expected pension benefits, personalizing the pension providers’ email communication is a first and crucial step. We describe a machine learning approach to model email newsletters to fit participants’ interests. The data for the modeling and analysis is collected from newsletters sent by a large Dutch pension provider of the Netherlands and is divided into two parts. The first part comprises 2,228,000 customers whereas the second part comprises the data of a pilot study, which took place in July 2018 with 465,711 participants. In both cases, our algorithm extracts features from continuous and categorical data using random forests, and then calculates node embeddings of the decision boundaries of the random forest. We illustrate the algorithm's effectiveness for the classification task, and how it can be used to perform data mining tasks. In order to confirm that the result is valid for more than one data set, we also illustrate the properties of our algorithm in benchmark data sets concerning churning. In the data sets considered, the proposed modeling demonstrates competitive performance with respect to other state of the art approaches based on random forests, achieving the best Area Under the Curve (AUC) in the pension data set (0.948). For the descriptive part, the algorithm can identify customer segmentations that can be used by marketing departments to better target their communication towards their customers. 








\end{abstract}
\section{Introduction}
\label{introduction}

Creating personalized content that is relevant and attractive to consumers has been one of the main challenges for marketers\cite{oberoi2017technology}. When consumers are flooded with text and link content emails, in the meanwhile, only a little or no content in the emails that is relevant to them. It leads to information crucial to their financial well-being escapes their attention; for instance, information about consumers’ pension plans which is pivotal for their long-term financial well-being. It is therefore of utmost importance that email communication is customized towards consumers’ interests.



Machine learning models offer to possibility of personalising marketing materials by better target newsletters and advertisements to the right population \cite{martinez2020machine}. The process typically involves collecting
data, cleaning it from outliers, reducing biases, and finally applying a machine learning algorithm to obtain an accurate prediction of the target variable and therefore calculate a conversion rate. The conversion rate can be any key performance indicator that matters for the model performance and business value. 


One of the main difficulties in machine learning is to identify the right features for the prediction task \cite{mckenzie1992construction}. A good model typically depends the selection of informative features. Deep learning (DL) apply an automatic feature engineering, which can result in good prediction outcomes but losing the intelligibility of the result, as the prediction relies on a non-linear combinations of the original features. Representation learning is one of the most successful approaches to extract complex features \cite{bengio2013representation}. The goal of representation learning is to identify good representations of the data, in order to obtain the best features for predictive tasks. Finding a low dimensional latent space in which to project the original variable a well.
Word embeddings are known representation learning approach \cite{mikolov2013distributed}. In their seminal work Mikolov et al. show that by means of shallow neural networks it is possible to obtain good vectorial representations of words and combinations of words by predicting their context within a corpus of documents of which the words are tokens. The natural extension of Mikolonov et al. has been the definition of embedding models for graphs \cite{grover2016node2vec, perozzi2014deepwalk} and knowledge graphs \cite{ristoski2016rdf2vec}. In graph and knowledge graph embeddings, the token to be represented becomes the node or concept in the graph, and the documents of the corpus to calculate the embeddings are simulated using random walks that navigate the graphs. Algorithms such as node2vec \cite{grover2016node2vec} and rdf2vec \cite{ristoski2016rdf2vec} have already shown great promise in modeling complex social network data, chemical processes \cite{wang2018prediction}, and bioinformatics knowledge \cite{10.1093/bioinformatics/btz718}.
The success of these algorithms, similarly to what happened with word embeddings, resides in the fact that embeddings calculated on top of a graph, retain information concerning the neighborhoods and connections taking place between the nodes, therefore enriching every single entity with semantic knowledge residing in the context of the entity.  

To successfully use graph embeddings of any sort, it is important to identify a graph structure that connects the entities in a meaningful way. Therefore, with respect to pure graph embeddings, the main challenge is to decide how two nodes are connected, the meaning of a connection and which entity should represent a node, a graph or a connection. 

This paper is, therefore, motivated by the problem of identifying an automated approach of generating such graphs when dealing with data that is both categorical and numerical. 

The main contribution of this paper is an approach, called random forests node embeddings (RFNE), to model categorical and continuous features by calculating node embeddings on the decision trees identified with a random forest algorithm. It shows that using embeddings calculated on a random forest is an effective method of extracting automated features for classification. It actually makes the classification more flexible as one can now use any algorithm in combination with random forests. The other significant contribution, is to show that by using the latent space features calculated with the embedding, we can come up with rules about a population of interest. This because we used a random forest to calculate the embeddings. The consequence is also that the nodes of the forest have a representation in the latent space. Therefore, if we take an interesting question as our dependent variable for a statistical test (i.e. subscribing to a loan or not) and the latent space features as the independent variables, we can then find those areas of the latent space in which the odds of finding the interesting target are higher. Once we find it, then we can find the closest node (nearest neighbour). Then closest node corresponds to a rule in the original feature space (because the node belongs to a decision tree, and we only need to navigate down the tree to find the rule).


To demonstrate the effectiveness of our approach, we use both open data and proprietary data to do the model evaluation. The open data comprises data sets about marketing and churn management, whereas the proprietary data belongs to a large pension fund of the Netherlands.

Such a contribution is significant to the marketing community because it allows to identify segments of the population in which the modelled advertisements are more effective. Our approach is not only use to calculate the predicted probability, but also it can be used to identify segmentation rules and provide advises for the marketers to select sub-groups. In addition, this contribution is also relevant to the machine learning community as it shows a new structured way of dealing with hidden spaces and embeddings and how to use the hidden space to further describe and explain the data for human consumption.


The rest of this paper is structured as follows: 
Section \ref{institutional} presents the motivating context and related work; Section \ref{methods} describes the main methods developed in this paper; Section \ref{evaluation} evaluates the model in four data sets: three open data sets for benchmarking the results (D1, D2, D3), and one private data set to strengthen the empirical evidence (D4). The results of these exercises are shown in Section \ref{evaluation}. Section \ref{conclusion} discusses the main implications of our work with respect to the marketing and machine learning disciplines, as well as lining out limitations and opportunities for future research.

\section{Context and Related Literature}
\label{institutional}

An environment in which marketers struggle to reach individuals is that of pensions. Around the world, people are reluctant to think about retirement; this results in a lack of adequate financially preparation \cite{binswanger2012real,wiener2008framework}. With regard to aging societies, this creates a major societal challenge with serious economic implications \cite{whiteford2006pension}. There have been many initiatives to encourage individuals to save for retirement, ranging from automatic enrollment \cite{beshears2010impact}, efforts to improve financial literacy (for a review see \cite{fernandes2014financial}) to improving the framing of communication \cite{behaghel2012framing}. 


\subsection{The Challenge of Engaging Consumers to Prepare for Retirement}
A challenge that pension providers face is communicating effectively with pension plan participants to foster learning and information gathering, and to increase awareness of the importance of pensions. The main goal is to engage participants to obtain information on their expected pension benefits and for them to determine whether they are on track with regards to their retirement savings. Pension providers in the Netherlands are obliged by law to provide their participants with information, and much of this communication happens online via emails and websites nowadays. However, engagement indicated by the clicking rate is low. Here the click rate is calculated by the number of participants who opened the email divided by the total number of participants who received the email.Personalizing an email by framing it in such a way that it fits the participants’ interests could potentially help to trigger participants to click. We assume that the content of an email is relevant to the participants' if they click the article links in a received email. Different article links lead participants to a different pension topic.

Past research has identified several barriers for participants to get informed about their pension. Making decisions for the far future i.e. intertemporal decision making, is difficult because participants are faced with more attractive realtime options: they can also spend their time watching a movie and or decide to spend their money on a new car rather than saving it for retirement. Searching for information takes time and effort, and retirement is so far away that participants postpone financial planning \cite{lynch2006you}. 

Participants may also be prone to information avoidance: while people know that information is available, and have free access to this information, they still decide to ignore it \cite{golman2017information}. A related barrier is information overload, meaning that when individuals are overwhelmed by the options available to them, they choose the path of least resistance \cite{agnew2005asset}, just as they do as a result of procrastination, status quo bias, and anticipated regret \cite{madrian2001power}. Choosing the path of least resistance can for example mean that individuals stick to the default, or do nothing like not reading the emails they receive from their pension provider. 

Given these barriers such as these, the question for pension providers is how to communicate with their participants in a way that these barriers are overcome. Pension providers have a duty to communicate and want to do this as efficiently as possible. Most people are already aware that saving for retirement is important, which makes them postpone engaging with it even more: important decisions take more time and effort, so people defer making them \cite{krijnen2015decision}. Communications on pensions that emphasize that it is important to save for retirement can, therefore, lead to opposite results.
Another way to communicate with participants is to personalize the communication they receive instead of sending a general message.

In this paper, we use an adaptive personalization approach to communicate with pension plan participants. In line with the definition of \cite{chung2016adaptive}, this means that personalization is done automatically based on algorithms, without the proactive efforts from the participant, and that it will adapt the product over time based on participant behavior. Adaptive personalization systems have been shown to be more successful than self-customization in news context \cite{chung2016adaptive}.

\subsection{Representation Learning and Embeddings Approaches}
\label{sec:relwork}

One of the fields that contributes the most towards representation learning is deep learning (DL) \cite{lecun2015deep}.
An important part of DL research is focused on modeling procedures to either learn approximate rules from data, like for example differentiable inductive logic programming \cite{evans2018learning} or to embed rules in a latent space in order to produce inference in the latent space, like for example visual query answering \cite{wang2018fvqa}, that embeds relationships amongst entities in pictures learning both the embedding of the entity and of the relationship and neural logic programming \cite{manhaeve2018deepproblog}, that instead applies rules to embeddings of complex objects to apply first-order logic reasoning. A similar approach is observable in the knowledge representation community, in which neural approaches are applied to learn relationships within an ontology \cite{hier2020neuro}, or neural networks are applied to perform approximated ontology based reasoning \cite{hohenecker2020ontology}.

Several contributions have previously made use of random forest as a structure to apply DL. Deep forest \cite{zhihua2019deepforest}, proposing a novel deep model multi-grained cascade forest (gcForest) model, a decision tree ensemble approach. The gcForest can have an outstanding performance on a small-scale data set with a fewer hyper-parameters than DL. 
Yuchuan Kong et al. \cite{yu2018deepnnm} proposed Forest Deep Neural Network (fDNN) algorithms to integrate the deep neural network architecture with a supervised forest feature detector. In the bioinformatics field, it is often the case that the number of predictors/features is much bigger than the number of instances and it limits the power of DL and the classical machine learning algorithms. The fDNN employed random forests as the feature detector, to learn sparse feature representations from a larger input feature space. 


Image Classification is one of the popular machine learning tasks in the image processing field. The images are represented by a set of independent patches characterized by local descriptors. Moosmann et al. \cite{FrankMNIPS2007} in 2007 used the Extremely Randomized Clustering Forests (ERC-Forests) to learn the spare local image descriptor. Each descriptor is represented by a set of leaf node indices with one element from each tree. They treated each leaf in each randomized decision tree as a separate visual word and stack the leaf indices from each decision tree into an extended code vector for each descriptor. In the paper \cite{FrankMNIPS2007}, the tree algorithm is used as a spatial partitioning method, to learn the representation of the image descriptor, not a classifier to label the target variable. The learned tree leaf features are the input features of the support vector machines (SVM) classifier for predicting the image.

Amongst the most successful recent representational approaches we can see that graph convolutional networks (GCN) \cite{zhang2018graph} are playing an important role in data structured in graph format. Differently from models based on random walks like node2vec, the GCN approach makes use of the adjacency matrix between the nodes of the graph, using convolutional filters to extract relationships between the nodes that can help performing a prediction task. GCN has been successfully used for link prediction \cite{schlichtkrull2018modeling}, text classification \cite{yao2019graph}, and spatio temporal tasks like traffic flow prediction \cite{guo2019attention} and activity recognition \cite{yan2018spatial}. With respect to the contribution presented in this paper, GCNs are currently being used to deal with data that already has a graphical representation, and similarly to random walk approaches, such as deep walk or node2vec, they could in principle be used to substitute the calculation of embeddings, given a method that extracts a graph representation from the data at hand. In the Limitations and Future work Section, we further discuss how these approaches could be used in synergy with the algorithm presented in this paper.

Another important related area of research concerns variational autoencoders (VAE) \cite{DBLP:journals/ftml/KingmaW19}. In general terms, autoencoders are generative models composed of an encoder and a decoder. The encoder is trained to project data items in a latent space, while the decoder is made to decode the data items from the latent space back to the original space.
VAEs extend the autoencoder model by specifying that the latent space variables should present a distribution (for example a Gaussian distribution). Given the flexibility of VAEs, many different variants of VAE have been defined, depending on the problem being addressed:
in \cite{ma2018constrained} Tengfei et al. defined VAE that can generate valid semantic expressions by means of a regularization procedure applied during the training of the VAE; in \cite{pu2016variational} Pu et al. use VAE to learn a latent representation of faces; in \cite{DBLP:conf/bigdataconf/LiaoGCL18} Liao et al. use VAEs to detect outliers; VAEs have also  been extended to deal with categorical variables \cite{DBLP:conf/iclr/JangGP17}.

With respect to the models presented in this section, the approach proposed in this contribution has the advantage of creating a latent space starting from a tree structured feature space. This implies that after performing inference and reasoning in the latent space, it is always possible to identify a rule in the original feature space. In addition to this, the proposed approach brings the advantage of being able to handle categorical data as a consequence of relying on decision trees for the embedding models, whereas the other models do not usually specify a strategy concerning such variables.




\section{Proposed Model}
\label{methods}

This section introduces the Random Forest Node Embeddings (RFNE) algorithm. RFNE relies on node2vec \cite{grover2016node2vec} for the calculation of node embeddings of decisions trees obtained by means of a random forest algorithm. Firstly we introduce the necessary background to comprehend RFNE, then we discuss RFNE producing examples to illustrate how the embeddings are calculated.

\subsection{Node Embeddings: the node2vec algorithm}
The main idea behind node embeddings frameworks is to learn continuous representations of nodes and edges in a network in order to be able to compare similar nodes in the network. The node2vec algorithm builds on top of ideas coming from natural language processing (word2Vec \cite{mikolov2013distributed}). In practical terms, node2vec treats each of the nodes like a word, the graph like a document composed of sentences created with random walks through the graph. Node2vec from Grover and Leskovec in 2016 \cite{grover2016node2vec} first applied a random walk over a graph to generate walk traces and extracts features based on a learned trace. How the trace moves through the graph is governed by a set of parameters:
\begin{itemize}
    \item $n$: number of walks
    \item $l$: length of the walk
    \item $p$: the return parameter $p$ measures the likelihood of immediately revisiting a node that has just been visited
    \item $q$: parameter $q$ differentiates between inward and outward nodes in the search.
\end{itemize}

Parameters $p$ and $q$ allow to differentiate between breadth first search (BFS) and depth first searchs (DFS). The random walk length $l$ and the number of random walks $n$ influence the neighbourhood taken into consideration and therefore the quality of the calculated embedding. 

\begin{figure}[h]
    \centering
    \includegraphics[scale=0.4]{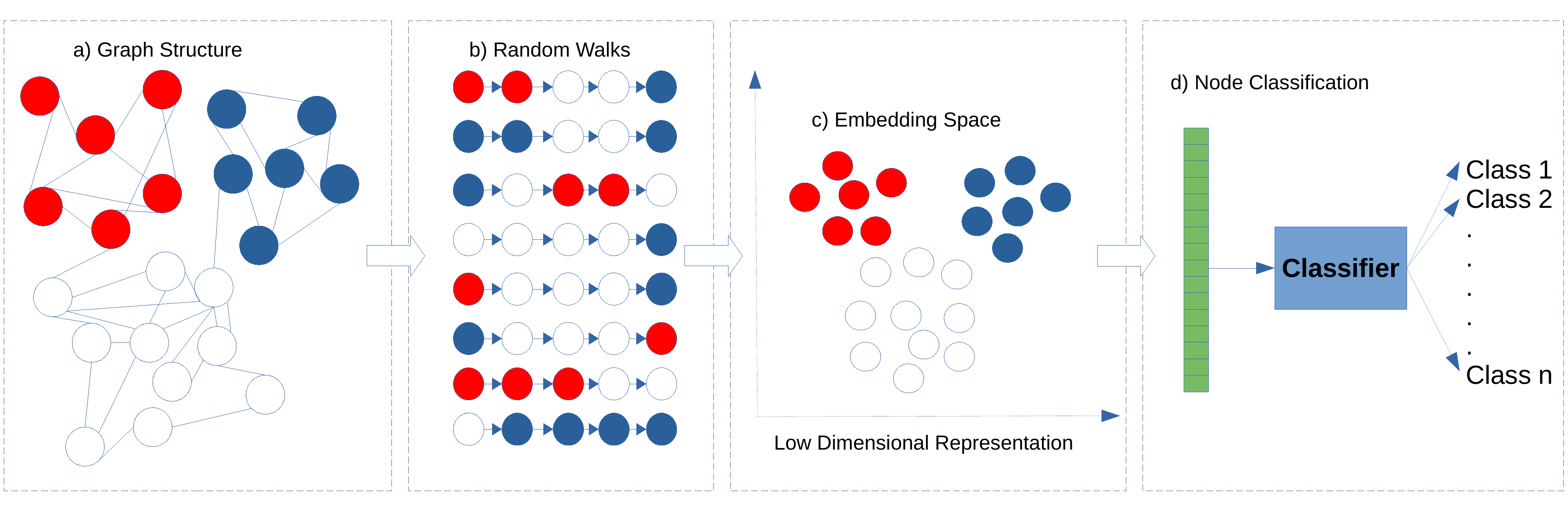}
    \caption{Illustration of node2vec process to calculate embeddings for node classification tasks.}
    \label{fig:n2vec}
\end{figure}

Figure \ref{fig:n2vec} illustrates the feature extraction process applied by the node2vec algorithm. In particular, after the random walks are calculated, node2vec only behaves as a wrapper for the word2vec algorithm, which will simply calculated the embeddings of the nodes as if they were words in a corpus of documents.

\subsection{Random Forests}
\label{RandomForests}

The Random Forest algorithm has been a very successful model for prediction in marketing contexts \cite{bart2015marketing} \cite{ja2017marketing}. It is an ensemble method that is based on training a large number of decision trees by means of different partitions of the training data (bagging) \cite{oza2005online}. 

\begin{figure}[h!]
    \centering
    \includegraphics[scale=0.3]{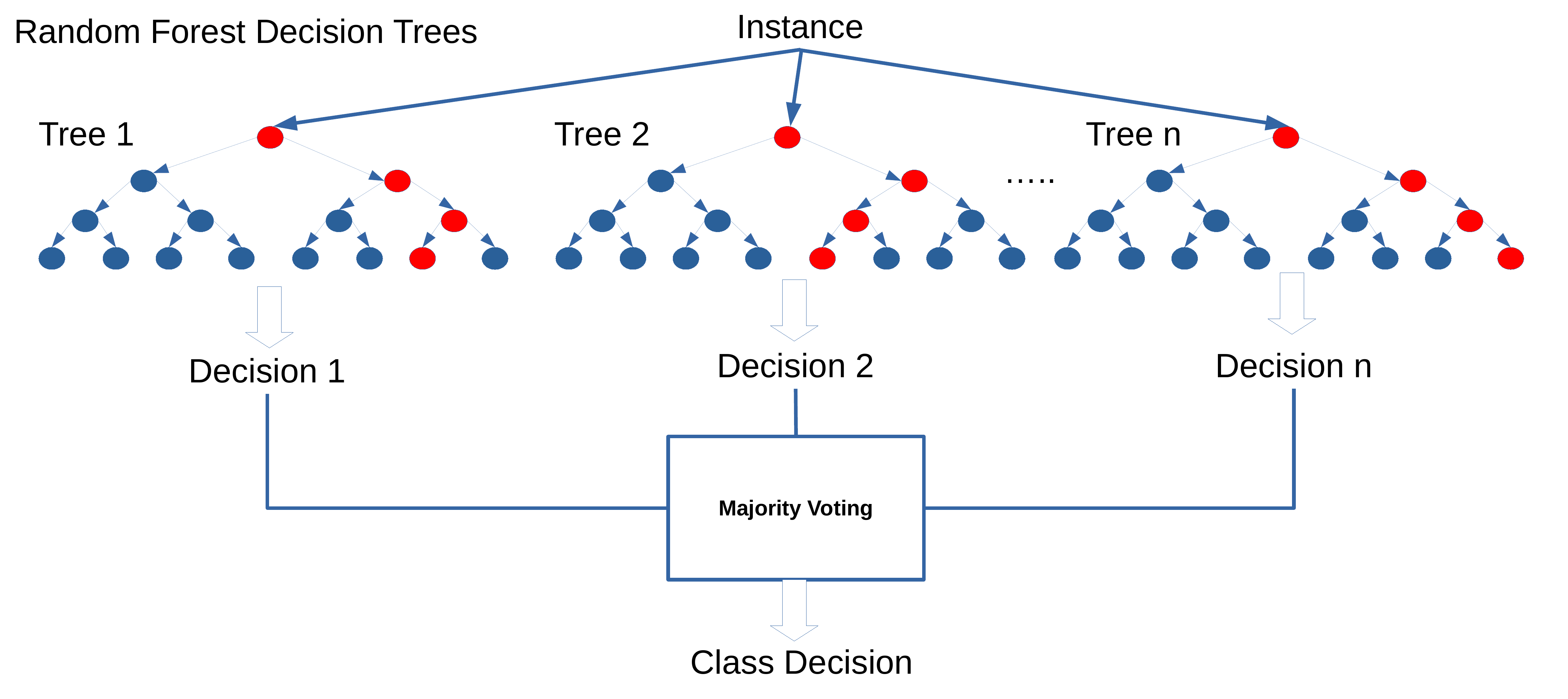}
    \caption{Random forests algorithm classification schema.}
    \label{fig:rfalgo}
\end{figure}
After the decision trees are trained, the final classification is performed by majority voting. Random forests are widely used in machine learning tasks because of their resilience against overfitting and their ability to generalize well even in complex data sets. One of the characteristics that makes this family of algorithms successful is that they can handle both continuous and categorical data. In addition to this, one of the main advantages of random forests with respect to other machine learning algorithms, such as SVMs \cite{suykens1999least}  and neural networks \cite{lecun2015deep}, is that random forests are interpretable thanks to the fact that the decision trees participating to the classification can be manually inspected.

\subsection{Random Forest Node Embeddings}

The algorithm developed in this contribution is shown in the Algorithm \ref{algo:alg1} pseudo-code snippet. The first step of the algorithm consists in calculating $k$ decision tree estimators with a random forest algorithm, and save them in set $T$. For the purpose of this contribution the random forest algorithm developed in \cite{scikit-learn0231} was used. 
The node2vec algorithm is then applied to each of the $t_i \in T$, with $|T| = k$ decision tree structures. The goal is to obtain $k$ embedding models $e_i \in E$ with $|E|=k$, that given a node identifier belonging to one of the trees can produce the embedding of an out-of-sample testing data record. 
The embedding features $f_j$ or each data record are calculated by using the leaves' identifiers $l_j$ for each of the decision trees considered. Each embedding model encodes information concerning the surrounding of the leaf.
Finally, embedding features $\tilde{x}_i$ represents the concatenation of the features calculated with all the embedding models and we concatenate it with $x_i$ features of the data record.

\begin{algorithm}[H]
\SetAlgoLined
 initialization\;
 $\mathbf{D} = \{ (x_1,y_1),(x_2,y_2),\ldots,(x_n,y_n) \}$\\
 $k$ estimators\\
 $l$ random walk length\\
 $r$ number of random walks\\
 $d$ embedding dimension\\
 $\tilde{D}$ $\leftarrow$ $\{\}$\\
 $E$ $\leftarrow$ $\{\}$\

 $ \mathbf{T} \leftarrow RandomForest(D,k)$ where $\mathbf{T} =\{ t_1,t_2,\ldots,t_k \} $ \
 
\ForAll{$t_i$ in $ \mathbf{T}$ with $t_i$ a decision tree structure }{
  instructions\;
  $ e_i$ $\leftarrow$ $node2vec(t_i,l,r)$ with $e_i$ and embedding model\\
  $E \leftarrow E \cup e_i$ with $E$ being an ensemble of embedding models\\
  
 }

 \ForAll{$(x_i,\_)$ in $D$}{
    \ForAll{$e_j$ in $E$}{
        $ l_j \leftarrow  calculate\_leaf(x_i,t_j) $ calculate node id in tree $t_j$ \\
        $f_j \leftarrow e_j(l_j) $ with $f_j$ in $\mathbb{R}^d$\\
        $\tilde{x_i}  \leftarrow \tilde{x_i} \cup f_j   $ forest embedding concatenation\\ 
        
    }
    
    $\tilde{D} \leftarrow \tilde{D} \cup (x_i \cup \tilde{x_i}, y)$ concatenate original features with calculated ones\\

 }
\KwResult{returns $\tilde{D}$ , $E$ }
\label{algo:alg1}
    \caption{Random Forest Node Embeddings Algorithm.}
\end{algorithm}

When such a feature extraction is applied, the embeddings of the leaves of the decision trees contain the information concerning the decision boundaries of the decision trees, including the neighbouring decision boundaries. The advantage of such a representation is that a linear discriminant model, such as logistic regression, can easily find a linear combination of the boundaries that would perform a good classification, allowing to substitute the majority voting of normal random forests with a soft weighing of the votes of the decision trees. In general terms, this approach can also be seen as a way of weighing the identified decision boundaries or as a way to select the most relevant decision trees. Algorithm \ref{algo:alg1} presents the case in which one combines the features by concatenating them together. This is not the only possible approach, another approach could involve taking a weighted average of the embeddings or to multiply the embeddings together. With respect to this paper, the focus has been on the concatenation of the features, because the concatenation allows us to perform complex descriptive reasoning, as discussed in Section \ref{evaluation}.

\begin{figure}[ht]
    \centering
    \includegraphics[scale=0.4]{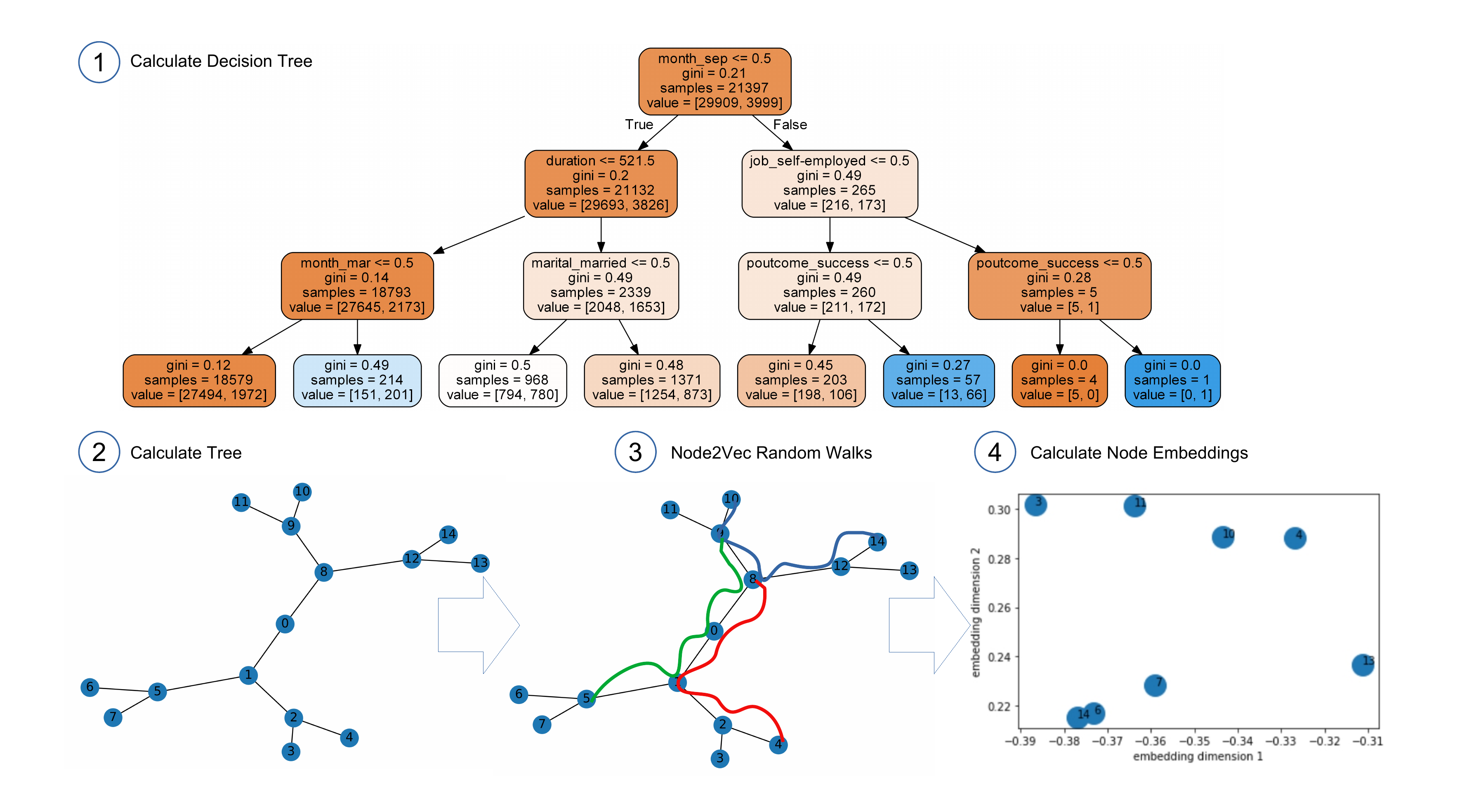}
    \caption{Calculating the node embeddings of a decision tree on the D1 benchmark data set.}
    \label{fig:embtree}
\end{figure}

The definition of the random walks length of the node2vec algorithm specifies the order of the neighbourhood of random forest embedding. Figure \ref{fig:embtree} shows a depiction of the embedding with a $l$ parameter equal to 15, on a tree calculated on top of the D1 data set used in this study. 

\section{Empirical Design}
\label{evaluation}

As a main case study to test our model, we use data of the largest pension fund in The Netherlands with over 2.8 million participants. Specifically, we use two data sets: one representing the general population participating in the fund, and the other one a sample in which a marketing campaign and an A/B testing experiment has been performed. In addition, we also tested our model in open data sets to confirm that the proposed model works also in different settings.

\subsection{Pension Fund Communication}

\begin{figure}[H]
    \centering
    \includegraphics[width=4in]{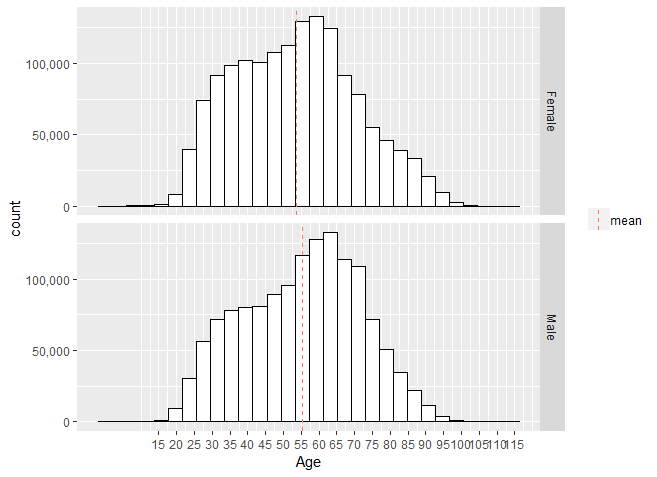}
    \caption{Overall fund population age by gender}
    \label{fig:ageABP}
\end{figure}

Employees working in the educational or governmental sectors are obliged to build up their pension at this fund. This results in a non-commercial duty for the fund (and by extension for their pension administrator) to provide their participants with information concerning their pension. One of the ways of communication with its participants is via an email newsletter, which they receive around four times a year. The newsletter is a collection of articles with information on five different categories: sustainability, choices regarding participants' pension, the financial situation of the fund, personal online page, and government decisions concerning pensions.
Age distribution by gender is shown in Figure \ref{fig:ageABP}, the mean age is 50.2 years. The total number of records in this data set is 2.8 millions. The features used in this study are shown in Table \ref{table:apg_table}.

\begin{figure}[h]
    \centering
    \includegraphics[width=4in]{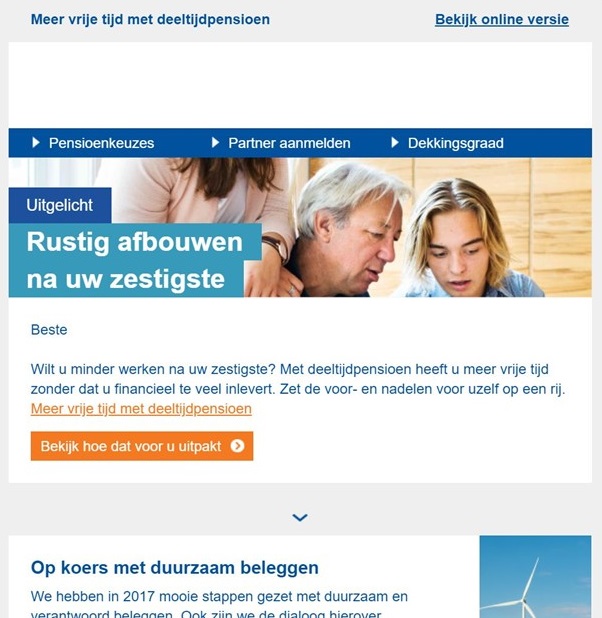}
    \caption{Example newsletter}
    \label{fig:screenshot}
\end{figure}

Figure \ref{fig:screenshot} shows a screenshot of one of the newsletters sent to the customers.

Since participants receive a lot of emails, sending only relevant information is crucial. To be able to provide these participants with useful information at the right time, individual participant profiles are created based on historical data.
The overall goal of these newsletters is twofold: first, to inform participants about important developments, and second, to get them more involved with their pension by having them click through on various articles that are interesting to them. 
In our field experiment, we focus on measuring and contributing to the second goal, achieving more pension involvement.

For the pension fund, it is important to have a model of the customers concerning who may be interested in a specific newsletter, to ensure maximum awareness in the customer population concerning pension topics. The data of the general population of the pension fund is denoted as D4 in the rest of the paper.

\begin{table}[H]
\resizebox{\textwidth}{!}{%
\begin{tabular}{|l|l|l|}
\hline
\textbf{Factor} &

  \textbf{Description} &
  \textbf{Type} \\ \hline
1. Participant Information  &
  \begin{tabular}[c]{@{}l@{}}Age in year\\ Marital status\\ Gender of a participant\\ Living province of a participant\\ Number of divorces of a participant\end{tabular} &
  \begin{tabular}[c]{@{}l@{}}Numerical\\ Categorical\\ Categorical\\ Categorical\\ Numerical\end{tabular} \\ \hline
2. Participant - Job &
   \begin{tabular}[c]{@{}l@{}}Sector of the last job\\ Number of funds to build up pension\\ Sector of the main job\\ Types of the main job regarding Dutch pension \\ rules, it includes normal retirement, earlier retirement, \\ unemployment benefit, flexible pension benefit\end{tabular} &
  \begin{tabular}[c]{@{}l@{}}Categorical\\ Categorical\\ Numerical\\ Categorical\end{tabular} \\ \hline
3. Participant - Financial &
  
  \begin{tabular}[c]{@{}l@{}}Actual year income\\ Part time factor of a job\\ Part time factor for the last job\\ Actual year income of the last job\end{tabular} &
  \begin{tabular}[c]{@{}l@{}}Numerical\\ Numerical\\ Numerical\\ Numerical\end{tabular} \\ \hline
4. Participant - Partner  &
  Partner at the same pension fund as the participant &
  Categorical \\ \hline
5. Communication choice  &
  The communication choice, digital or letter &
  Categorical \\ \hline
6. Pension Product &
  Does a participant buy the specific pension product? &
  Categorical \\ \hline
\end{tabular}%
}
\caption{Feature description for the Dutch Pension Fund data.}
\label{table:apg_table}
\end{table}

\subsection{Pilot Sample Characteristics}

In addition to the general population data of the pension fund, we also use data from a field experiment that took place in July 2018. The experimental hypothesis of the experiment is that  the click rate of the intervention group is higher than that of the control group, given the application of a machine learning model to target the intervention group. 

All $463,212$ active (i.e. accruing pension) and subscribed pension plan participants at the fund received a newsletter. 52\% of the sample are male, the mean age is 48.4 years for females and 52.1 years for males, and the mean annual income is 50,317.11 Euro.

\begin{figure}[H]
    \centering
    \includegraphics[width=4in]{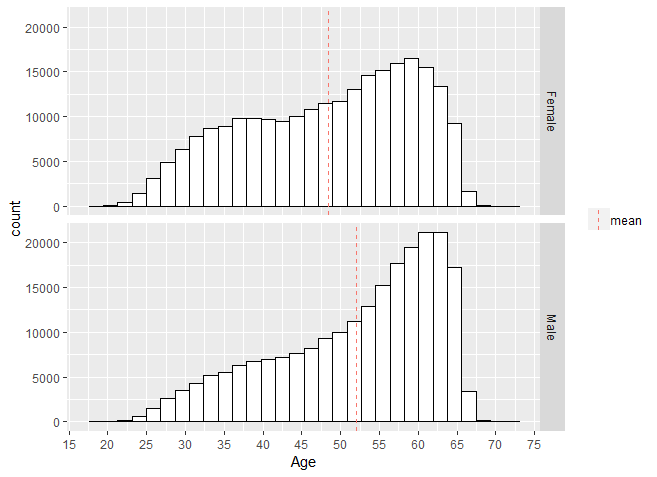}
    \caption{Pilot: population age by gender}
    \label{fig:agepilot}
\end{figure}

Figure \ref{fig:agepilot} shows the distribution of sample age by gender.
With regards to participants' marital status, 62.2\% are married, 16.9\% are single, 10.5\% have a registered partner, and 10.4\% are divorced. 72\% of the sample work fulltime (that is, more than 0.75 fte), the rest work parttime (that is, less than 0.75 fte).

The features used for this study are the same as those reported in Table \ref{table:apg_table}.

In this paper, we chose the Pension Choice topic as an example for further analysis. There are in total of $227,885$ participants involved in such a topic. The intervention and control groups have $89,587$ and $138,298$ participants respectively. For analysis of the pilot results, it is important to note that each click on an article gets stored at an individual level. Therefore, we can extract which participants (by their IDs) clicked on which article at what moment in time. 

\begin{table}[H]
\centering
\begin{tabular}{|l|c|c|}
\hline
\textbf{Group} & \multicolumn{1}{l|}{\textbf{Experimental Group}} & \multicolumn{1}{l|}{\textbf{Control Group}} \\ \hline
\textbf{Pension Topic Articles} & Pension Choice & Pension Choice  \\ \hline
\textbf{Selected Participants}  & The Pre-trainned Model  & Random  \\ \hline
\textbf{Number of participants} & 89,587         & 138,298         \\ \hline
\end{tabular}%
\caption{The Pilot design: the pension fund sent the same email with the Pension Choice Articles to participants in both intervention and control group. The only difference between the two groups is that the participants from the intervention group are selected by the pre-trained model.}
\label{tab:pilot-table}
\end{table}
The Pre-trained model is a binary logistic regression model (BLR), which is used to predict the interesting topic (Pension Choice) of an individual participant. Firstly, this model is applied to the overall data set to select the participants most likely to be interested in the Pension Choice topic to construct the intervention group. The selected participants are eliminated from the overall data set. After that, we randomly selected 138,298 participants from the remaining data set without replacement to construct the control group. 

The click rate of the participants from the intervention and control groups is $29.39\%$, and  $16.91\%$ respectively, based on the pilot data. The $Chi^2$ test (p-value $<0.01$) shows that the click rate of the Experimental Group is significantly higher than that of the Control Group. The current BLR model can help the pension fund to predict the interesting topic of an individual participant. The findings show that pension fund can use their historical data on participant behaviour to improve their communication. 

In the study performed, the data collected showed that the BLR model works, given the large experiment sample. In reality, the budget available for campaigns is often limited to smaller subsets. In this paper we therefore show how RFNE can help selecting subsegments of the population in which the advertisement would be particularly effective.


\subsection{Benchmark Data}

Due to the data protection regulation of pension data in the Netherlands, we cannot publish our research pension data. To help the other researchers reproduce the current results and continue our research, we chose a number of publicly available data sets concerning problems of churning and  marketing campaigns.
First of all we use the Portuguese banking institution data (\cite{benchmarkdataset}, denoted as D1 in the rest of the paper). The data is collected from $41,188$ bank clients from  May 2008 to November 2010 and published in February 2012. The marketing campaigns were based on phone calls. The data has been preprocessed by the original authors, and it includes 20 features in total. 
The data set target reports the success in convincing a is a client to subscribe a bank term deposit.

Secondly, we make use of the KDD 2009 challenge (\cite{DBLP:journals/sigkdd/GuyonLBDV09}, denoted as D2 in the rest of the paper) that provides the data set of a large telecommunication company (Orange) to predict the propensity of a customer to switch to another provider, comprising $227$ features, of which 38 of categorical nature, $189$ of numerical nature, plus one target variable reporting churning and non churning customers ($3672$ and $46328$ customers respectively). '

Finally, we also make use of the Duke data set (denoted as D3 in the rest of the paper), that concerns $49752$ data records, containing $56$ variables, of which $22$ categorical features, $33$ numerical features and $1$ target variable reporting churning and non churning customers.

\subsection{Results}

We perform our evaluation in the two use cases by splitting our data 80\% training and 20\% testing.

To evaluate the performance of the algorithm, we performed a grid search using a 5-fold cross validation on the training data on a number of parameters. For a matter of comparison, the parameters of the machine learning models identified on the standard data without embeddings, are also re-used for the part with node2vec embeddings. Table \ref{table:parameter_table} shows the parameters used for the grid search and the selected values.

\begin{table}[H]
\resizebox{\textwidth}{!}{%
\begin{tabular}{l l l l l l}
\hline
Algorithm   & parameters Range    & Parameters D1     & Parameters D2     & Parameters D3  & Parameters D4      \\ \hline
KNN &
  k $\in$ {[}1,10{]} step 1 &
  3 &
  5 &
  10 &
  5 \\ \hline
RandomForest &
  \begin{tabular}[c]{@{}l@{}}n $\in$ {[}10,200{]} step 10\\ levels  $\in$ {[}2 to 20{]} step 1\end{tabular} &
  \begin{tabular}[c]{@{}l@{}}200\\ 5\end{tabular} &
  \begin{tabular}[c]{@{}l@{}}150\\ 10\end{tabular} &
  \begin{tabular}[c]{@{}l@{}}200\\ 10\end{tabular} &
  \begin{tabular}[c]{@{}l@{}}200\\ 10\end{tabular} \\ \hline
ExtraTrees &
  \begin{tabular}[c]{@{}l@{}}n $\in$ {[}10,200{]} step 10\\ levels $\in$ {[}2 to 20{]} step 1\end{tabular} &
  \begin{tabular}[c]{@{}l@{}}200\\ 5\end{tabular} &
  \begin{tabular}[c]{@{}l@{}}150\\ 10\end{tabular} &
  \begin{tabular}[c]{@{}l@{}}200\\ 10\end{tabular} &
  \begin{tabular}[c]{@{}l@{}}200\\ 10\end{tabular} \\ \hline
  
Logistic Regression &
  \begin{tabular}[c]{@{}l@{}}Regularization in \{'None', 'L1', 'L2'\}\\ max iterations $\in$ {[}100,500{]} step 100\end{tabular} &
  L2 &
  L2 &
  L2 &
  L2 \\ \hline                                                     \\
\begin{tabular}[c]{@{}l@{}}RandomForest \\ Node Embeddings\end{tabular} &
  \begin{tabular}[c]{@{}l@{}}l $\in$ \{5,10,15\} step 5\\ r $\in$ {[}20,100{]} step 20\\ d $\in$ {[}1,5{]} step 1\end{tabular} &
  \begin{tabular}[c]{@{}l@{}}5\\ 50\\ 10\end{tabular} &
  \begin{tabular}[c]{@{}l@{}}5\\ 100 \\ 10\end{tabular} &
  \begin{tabular}[c]{@{}l@{}}5\\ 100\\ 50\end{tabular} &
  \begin{tabular}[c]{@{}l@{}}5\\ 100\\ 50\end{tabular}  \\ \hline
\end{tabular}%
}
\caption{Parameters used to optimize the algorithms.} 
\label{table:parameter_table}
\end{table}

\begin{table}[H]
\begin{center}
    
\resizebox{0.8\textwidth}{!}{%
\begin{tabular}{l l l l l}
\hline
\textbf{Method}       & \textbf{D1} & \textbf{D2} & \textbf{D3} & \textbf{D4} \\ \hline
KNN                   &      0.849        & 0.533         & 0.566           & 0.931\\ \hline
RandomForest          &     {\bf 0.922}   & {\bf 0.657}   & 0.707           & 0.947\\ \hline

ExtraTrees            &     0.920         & 0.624         & 0.686           & 0.945\\ \hline
Logistic Regression   &     0.861         &   0.608       & 0.641           & 0.941\\ \hline
Random Forest Node Embeddings & 0.907     & {\bf 0.657}   & {\bf 0.722}     & {\bf 0.948} \\ \hline
\end{tabular}%
}
\end{center}

\caption{AUC score comparison of the selected algorithms in the studied data sets. The best performers are shown in bold.}
\label{tab:my-table}
\end{table}

As shown in \ref{tab:my-table} in the selected data sets RFNE presents a competitive performance, from the perspective of classifying data sets presenting both categorical and numerical features.
An interesting aspect of this evaluation is that it suggests that embeddings built on the decision trees identified by the random forest algorithm retain the information of the original data and that the combination of such embeddings with a logistic regression algorithm allows to produce an improvement with respect to a random forest algorithm. This improvement can be explained by the fact that the logistic regression acts applied on the embeddings acts as a soft voting mechanism that in addition includes an optimization of the vote weights based on the LBFGS solver \cite{liu1989algorithms}, where instead a basic random forest applies majority voting. 
The comparison performed above covers only one part of the possibilities of extending RFNE. For example, given the sequence of embeddings representing the decision trees of the random forest, any algorithm which is able to work with sequences could be used to identify relationships between the trees useful for the sequence classification process, we leave these and other investigations to future works and discuss them in Section \ref{conclusion}.




\subsection{Random Forests Node Embeddings As a Description and Verification Tool}

The current practice towards mining relevant patterns by using random forests implies a linear search amongst the leaves of the trees composing the forest, performing comparisons in terms of frequencies of the entities of interest. As the potential leaves in a big data set can be many, the process can turn out to be rather time demanding. In addition, comparing multiple variables and multiple groups identified by the trees in the forest implies performing many set operations, that are typically very demanding from a computational perspective.

Since RFNE is a technique stemming from random forests, the hidden space identified with node2vec can be associated with the trees and that can be further used to perform statistical tests to identify customer segments of interest.
In specific terms, by using RFNE, this discovery process can be reduced to a statistical test between the embedded variables and a dependent variable, followed by a nearest neighbour search in the hidden space to find the closest leaf to the center of mass (i.e. the mean vector) of a population of interest. Remembering that each leaf has also a partitioning rule associated, once the leaf or leaves are identified in the hidden space, then a potentially highly descriptive set of rules for the dependent variable can also be obtained.

\begin{figure}
    \centering
    \includegraphics[scale=0.3]{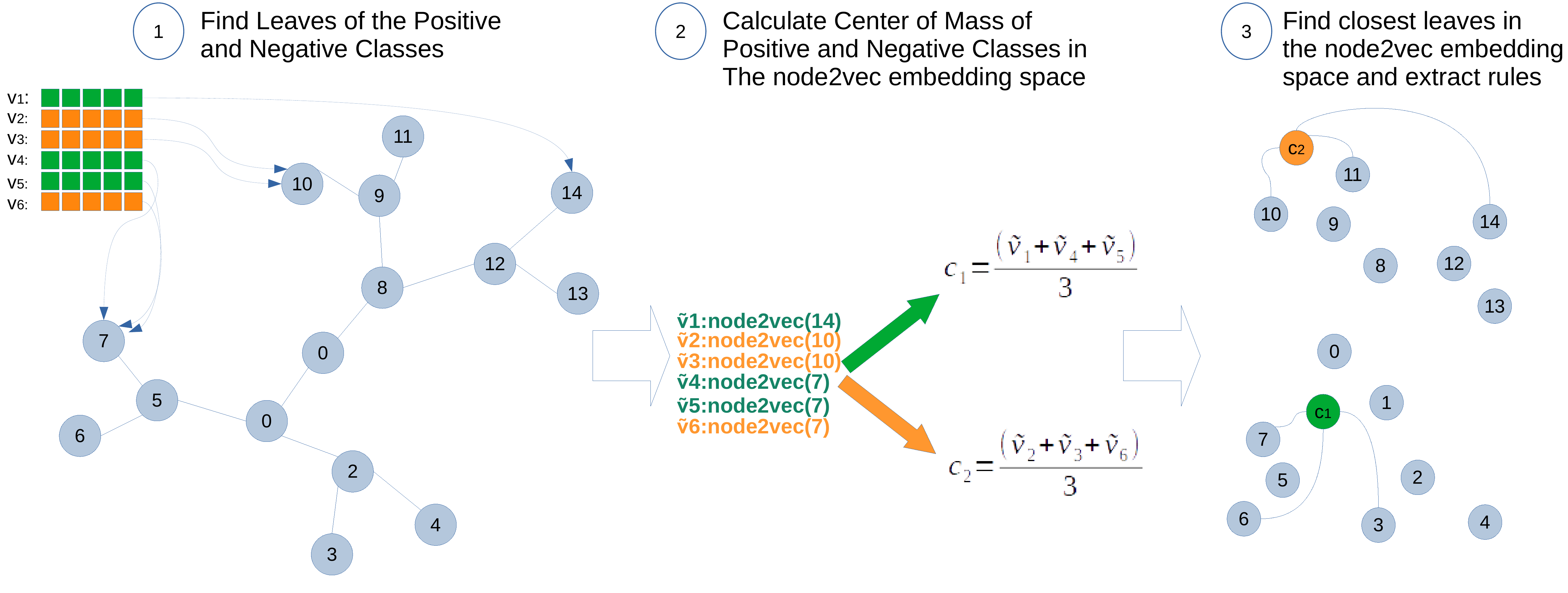}
    \caption{Rule Extraction Process.}
    \label{fig:ruleextract}
\end{figure}

Fig. \ref{fig:ruleextract} shows a depiction of the rule extraction process.

In the following description, we will use the D1 data set, the general pension fund data, and the pilot pension fund data. Each of these studies has been carried out by setting $10$ decision trees, with maximum $5$ levels of depth, with each of the nodes presenting $50$ random walks of dimension $10$.

\subsection{Rule Discovery in the D1 Data Set}
In the D1 data set, there are several variables that present categorical nature. We will focus on using binary variables as our dependent variables of study, so as to use widely recognized statistical tests such as a logistic regression. D1 data set concerns bank data and we  perform a logistic regression using {\bf loan} as the target variable to identify segments of the population that would subscribe to a loan. 



\begin{table}
\begin{center}
\begin{tabular}{lrrrrrrr}
\hline
{} &  log odds &   stde &       z &  P$>|z|$ &  $[$0.025 &  0.975$]$ &      odds \\
\hline
Intercept  &   -4.0580 &  0.186 & -21.846 &  0.000 &  -4.422 & -3.694 &  0.017284 \\
feature\_0  &    0.2344 &  0.134 &   1.751 &  0.080 &  -0.028 &  0.497 &  1.264150 \\
feature\_1  &    0.2812 &  0.078 &   3.620 &  0.000 &   0.129 &  0.433 &  1.324719 \\
{\bf feature\_2}  &    0.3986 &  0.067 &   5.967 &  0.000 &   0.268 &  0.530 &  1.489738 \\
{\bf feature\_3}  &    0.7227 &  0.098 &   7.346 &  0.000 &   0.530 &  0.916 &  2.059988 \\
feature\_4  &   -0.2204 &  0.069 &  -3.201 &  0.001 &  -0.355 & -0.085 &  0.802198 \\
feature\_5  &    0.0864 &  0.093 &   0.924 &  0.355 &  -0.097 &  0.270 &  1.090242 \\
feature\_6  &   -0.1506 &  0.108 &  -1.392 &  0.164 &  -0.362 &  0.061 &  0.860192 \\
feature\_7  &   -0.0289 &  0.076 &  -0.378 &  0.706 &  -0.179 &  0.121 &  0.971514 \\
feature\_8  &   -0.1441 &  0.098 &  -1.473 &  0.141 &  -0.336 &  0.048 &  0.865801 \\
feature\_9  &    0.6914 &  0.096 &   7.192 &  0.000 &   0.503 &  0.880 &  1.996509 \\
feature\_10 &   -0.2434 &  0.133 &  -1.824 &  0.068 &  -0.505 &  0.018 &  0.783958 \\
feature\_11 &    1.2346 &  0.132 &   9.320 &  0.000 &   0.975 &  1.494 &  3.437003 \\
feature\_12 &   -0.7970 &  0.121 &  -6.560 &  0.000 &  -1.035 & -0.559 &  0.450679 \\
feature\_13 &   -0.3707 &  0.134 &  -2.776 &  0.006 &  -0.632 & -0.109 &  0.690251 \\
feature\_14 &   -0.4318 &  0.122 &  -3.554 &  0.000 &  -0.670 & -0.194 &  0.649339 \\
feature\_15 &    0.2506 &  0.115 &   2.187 &  0.029 &   0.026 &  0.475 &  1.284796 \\
feature\_16 &    0.0913 &  0.117 &   0.782 &  0.434 &  -0.138 &  0.320 &  1.095598 \\
feature\_17 &   -0.1963 &  0.076 &  -2.567 &  0.010 &  -0.346 & -0.046 &  0.821766 \\
feature\_18 &   -0.2055 &  0.168 &  -1.223 &  0.221 &  -0.535 &  0.124 &  0.814240 \\
feature\_19 &   -0.2724 &  0.077 &  -3.559 &  0.000 &  -0.422 & -0.122 &  0.761550 \\
\end{tabular}
\end{center}
\caption{RFNE Features effect on the dependent variable 'loan' in the D1 data set.}
\label{tab:statsportu}
\end{table}

In Table \ref{tab:statsportu} we find that feature $2$ and $3$, belonging to the $2^{nd}$ tree of the feature extraction would have a positive log-odds towards the dependent variable. We can identify the mean vector of the entities in our database with respect to these features and then compare it, in terms of euclidean distance, with respect to the leaves of the $2^{nd}$ tree. This enables the identification of the closest leave, that also corresponds to the following rules with respect to the D1 data set:

\begin{lstlisting}[mathescape=true]
$rule_{d1}$ =pdays < 9.5 and  marital != 'single' and 
           job $\in$ ["admin.", "blue-collar", "entrepreneur", "housemaid"] and
           age < 61.5 and day > 18.5
\end{lstlisting}

where $pday$ represents the number of days after the last contact with the customer, and $day$ represents the day in the month. This can identify a segment in which the concentration of people who subscribed to a loan would be higher than the rest of the population (25.2\% vs 18.4\% in the rest of the population). It is also possible to check if this population would have a higher probability of defaulting, that is the original target variable of the D1 data set (13.3\% vs 10\% in the rest of the population). The value of identifying $rule_{d1}$ with respect to D1 is that it provides a segment of the population in which the bank would have more risk than with the rest of the customers, as at the same time the customer in this segment would have a higher probability of defaulting while also having a loan with the bank at the same time.


\subsection{Rule Discovery in the Pension Pilot Data Set}



The approach described in this section can also be used to identify relevant segments in A/B testing trials. Table \ref{tab:log_odds} shows the relationship between the features calculated with RFNE and a target variable discriminating between intervention and control groups. As all the features seem to be significant with respect to the target variable, in order to identify relevant patterns we select those features that present the best odds concerning finding a relevant difference between intervention and control, that in this case are feature\_2 and feature\_3 that are associated with the first tree.

\begin{table}
\begin{center}
\begin{tabular}{lrrrrrrr}

{} &    log odds &  std err &       z &  $P>|z|$ &  [0.025 &  0.975] &  odds \\
Intercept  & -1.8638 &    0.113 & -16.485 &    0.0 &  -2.085 &  -1.642 &   0.155082 \\
{\bf feature\_0}  &  1.1599 &    0.015 &  79.510 &    0.0 &   1.131 &   1.189 &   3.189614 \\
{\bf feature\_1}  &  2.6814 &    0.063 &  42.407 &    0.0 &   2.557 &   2.805 &  14.605527 \\
\end{tabular}
\end{center}
\caption{RFNE Features effect on the dependent variable 'Intervention' in the APG pension trial.}
\label{tab:log_odds}
\end{table}

By selecting those features and applying the approximated nearest neighbour procedure illustrated in Fig. \ref{fig:ruleextract}
we can identify the following rule.

\begin{lstlisting}[mathescape=true]
$rule_1$ = newsletter = 'yes' and  age > 60.6 
           part_time_factor <= 0.94, part_time_factor >=0.48
\end{lstlisting}

As we have intervention and control groups, we can observe if this segment presents statistically significant differences with respect to the clicking behaviour by using a $Chi^2$ test. This is illustrated in Table \ref{tab:rule1tab}. 

\begin{table}[H]
\centering
\begin{tabular}{@{}p{2cm}p{3cm}p{3cm}@{}}
\hline
 & Control & Intervention \\ 
\hline
Click           & 5764/130000                & 5517/90000                      \\
No Click         & 20/130000                  & 3551/90000                      \\
& \multicolumn{2}{l}{p-value $<$ 0.01}                                        \\ \hline
\end{tabular}
\caption{Contingency Table and p-value of $Chi^2$ test for Control and Intervention, after applying rule 1.}
\label{tab:rule1tab}
\end{table}

With respect to the leaf in consideration, the $Chi^2$ test shows that the intervention population presents a different behaviour with respect to the process of clicking the newsletter (p-value $<0.01$).

In the case in which we wanted to further restrict our population in consideration, one could define an additional dependent variable and identify another tree in the forest to produce additional rules. For example if we perform this operation for the gender variable in the pension data set, we can obtain another table of effects in which some of the trees show a big effect with respect to gender and we can also extract the additional rule below:

\begin{lstlisting}[mathescape=true]
$rule_2$ = gender = 'woman' and  
           main_job in ['academic hospital','waste collector', 
                        'education', 'defence','energy','police']
           and part_time_factor <= 0.71 and associate_partner = 'yes'
\end{lstlisting}

which, combined with the first rule identified, produces the following contingency table.

\begin{table}[H]
\centering
\begin{tabular}{@{}p{2cm}p{3cm}p{3cm}@{}}
\hline
 & Control & Intervention \\ 
\hline
Click           & 420/130000                & 412/90000                      \\
No Click         & 1/130000                  & 254/90000                      \\
& \multicolumn{2}{l}{p-value $<$ 0.001 }                                        \\ \hline
\end{tabular}
\caption{Contingency Table and p-value of $Chi^2$ test for Control and Intervention, after applying rule 1 and rule 2.}
\end{table}

These results suggest customers of the pension fund in which the applied intervention predicts a difference in the clicking behaviour. Specifically, concerning the first rule, RFNE shows that it is possible to identify areas of the hidden space in which the density of the population clicking the newsletter is higher ($6.1\% vs 4.4\%$), but that also implies more false positives ($4\% vs 0.1\%$) when using the selected intervention. When the second rule is applied, we can focus on a sub population based on gender, seeing a similar result in terms of false positives and false negatives.

\section{Discussion}
\label{conclusion}

\subsection{Implications for Marketing}

In this paper, we developed an adaptive personalization model to get pension plan participants more involved with their pension. We conducted a field experiment in which we sent out a newsletter to all active 465,711 participants of a pension fund, and find that our model results in significantly higher clicking rates. 


Irrelevant communication can annoy participants with potential detrimental effects, such as having them unsubscribing from the mailing list \cite{damgaard2018hidden}. Individuals receive a lot of emails every day from various service providers.
Not all emails are as important as the ones from their pension fund, so pension providers should help participants view their communication as relevant by optimizing their mailing strategy.

In addition, it is important to reduce money management stress and negativity around financial planning, factors which have shown to negatively influence financial and overall well-being \cite{netemeyer2017doing}. Our findings show that pension funds can use their historic data on participant behaviour to predict what topics from the email they will select, and potentially improving the effectiveness of customized email, and potentially yielding better informed pension savers and improved pension saving outcomes.


\subsection{Implication for the Machine Learning Community}
Node embeddings have already had an impact in changing the approach towards managing data and considering relationships between entities and features that follow a non-traditional schema. In the traditional machine learning schema, the feature engineering process ended after the first definition of the features. This contribution within a stream of research that reexamines the typical assumptions of machine learning algorithms, such as having samples that are identically distributed and independently sampled (iid assumption): the very fact that a node2vec representation of the decision areas in the random forest improves the results, implies that the iid assumption is very strong in many cases, as that the proximity of decision regions, and therefore of the entities falling into such regions, influences the decision boundary. Recent contributions in relational and representational learning have shown that relaxing the iid assumption can lead to better results in many situations.
Our contribution opens opportunities of further development. We applied random forests without performing further transformations to the data. Many more opportunities exist, including:

\begin{itemize}
    \item Combining random forest with clustering approaches: applying clustering before using the random forest may lead to a different partitioning of the data that could yield better  by uncovering partitions in the data due to different densities and groupings.
    
    \item Substitute random forest with rule induction approaches: advanced association rule mining algorithms, such as FP-GROWTH, make use of tree structures to store the transactions, those structures could be used to calculate embeddings for each of the transactions to be classified.
    \item Use the the random forest embeddingss as features for regression and autoregression: in the current contribution the random forest is used for classification purposes, but the same approach could be effective with trends and time series predictions.
\end{itemize}

While this paper has focused on the node2vec technique to extract the embeddings, other techniques exist to extract node, graph and subgraph embeddings that alone or in combination may be interesting to explore future.

This paper describes a machine learning approach to identify customers that would be receptive to a particular newsletter.We discussed an approach, based on random forests and node embeddings, to find relationships within features that can enhance the precision and recall of the classification process. The algorithm works by first calculating an ensemble of decision trees, then calculating a graph connecting the leaves of the decision trees and finally by applying node2vec on the calculated graphs to extract the node embeddings of the leaves. The concatenation of the node embeddings is then used as features for the classification of the customers with respect to a newsletter topic.

\subsection{Limitations and Future Work}
Our research is not without limitations. 
while we can predict with reasonable accuracy which topics of a email are selected by an individual, we have no additional knowledge as to whether doing so has any impact on customer satisfaction or on actual pension savings. Further interventional studies will be required to validate the utility of our approach.



Second, it would be interesting to know what role timing and frequency of emails play, and whether these would need to be adapted for different individuals as well.


The proposed approach comes at the additional cost of calculating node embeddings on the structure of the decision trees, in order to uncover the hidden space that allows us to perform further reasoning. As such, RFNE cannot scale as well as random forests. Nonetheless, the pension fund data set studied in this paper contained 2 million entries, meaning that, despite the additional computational burden of node2vec, RFNE can scale up to fairly large data sets. Concerning the descriptive part, the current approach focused on uncovering patterns occurring between binary variables.

We envision several studies that could follow this one. In this contribution we used node embeddings to find relationships between categorical and continuous features of customer data. An interesting future direction implies making use of knowledge graphs and ontologies to calculate the embeddings and then compare the results with the embeddings calculated on the random forest trees. In the paper, the graph is constructed by a collection of decision trees. A decision tree is generated based on the attribution selection measures, such as Information Gain, Gain Ration, and Gini Index. While a knowledge graph represents a collection of interlinked descriptions of entities/features. It puts data in context via linked and semantic. Therefore, the embeddings are calculated from the knowledge graph carries the nature relationships among features compare to its from random forest trees. Another promising direction is to consider different types of feature extractors rather than random forests and decision trees, as for example using a rule induction engine \cite{rules_from_data} and then embed the tree of the rules rather than decision trees as in this contribution. 
Finally, the feature extraction presented in this paper offers opportunities in the direction of generative models. Since the embeddings are calculated from regions identified by decision trees, when coupled with variational autoencoders \cite{zhang2019d} to reconstruct the original signal, they could potentially create realistic reconstructions, allowing to create a powerful tool for data anonymization.

\printbibliography
\end{document}